%% file: main.tex
\documentclass{article} 
\usepackage{iclr2024_conference,times}
\iclrfinalcopy 

\input{math_commands.tex}

\usepackage{hyperref}
\usepackage{url}
\usepackage{booktabs}
\usepackage{multirow}

\usepackage{tikz}
\usetikzlibrary{automata, positioning, arrows.meta, shapes.geometric}

\usepackage{algorithm}
\usepackage[noend]{algpseudocode}
\usepackage{listings}
\usepackage{xcolor}  

\lstdefinestyle{mypy}{
    language=Python,
    basicstyle=\ttfamily\scriptsize,
    keywordstyle=\color{blue},
    commentstyle=\color{green!50!black},
    stringstyle=\color{red},
    showstringspaces=false,
    breaklines=true,
    frame=single,
    rulecolor=\color{gray},
    framesep=3pt,
    frameround=ffff,
    fillcolor=\color{white},
    backgroundcolor=\color{white}
}

\usepackage{xspace}

\newcommand{\rest}{\texttt{ReST}\xspace}
\newcommand{\palmL}{\texttt{PaLM 2-L}\xspace}
\newcommand{\palmS}{\texttt{PaLM 2-S}\xspace}
\newcommand{\palmXS}{\texttt{PaLM 2-XS}\xspace}

\usepackage{graphicx}
\usepackage{wrapfig}
\usepackage[labelfont=bf, font=small, textfont={it,sf}]{caption}
\usepackage{xcolor}
\usepackage{authblk}

\title{ReST meets ReAct: Self-Improvement for Multi-Step Reasoning LLM Agent}

\author[$\dagger$1]{\bf Renat~Aksitov}
\author[$\dagger$1]{\bf Sobhan~Miryoosefi}
\author[$\dagger$1]{\bf Zonglin~Li}
\author[$\dagger$1]{\bf Daliang~Li}
\author[$\dagger$2]{\bf Sheila~Babayan}
\author[$\dagger$2]{\bf Kavya~Kopparapu}
\author[1]{\bf Zachary~Fisher}
\author[1]{\bf Ruiqi~Guo}
\author[1]{\bf Sushant~Prakash}
\author[3]{\bf Pranesh~Srinivasan}
\author[2]{\bf Manzil~Zaheer}
\author[1]{\bf Felix~Yu}
\author[1]{\bf Sanjiv~Kumar}

\affil[ ]{\small $^1$Google Research, $^2$Google DeepMind, $^3$Google}
\affil[ ]{\small $^\dagger$Core contributors}

\renewcommand{\headrulewidth}{1pt}
\fancypagestyle{firstpage}{
  \lhead{\begin{picture}(0,0)\put(0,-3){\includegraphics[width=0.25\linewidth]{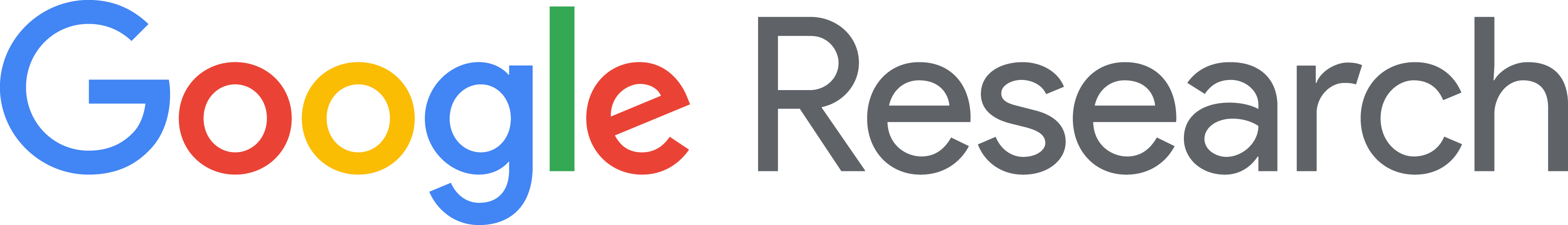}}\end{picture}}
}

\begin{document}

\maketitle

\input{macros}

\thispagestyle{firstpage}

\begin{abstract}

Answering complex natural language questions often necessitates multi-step reasoning and integrating external information. Several systems have combined knowledge retrieval with a large language model (LLM) to answer such questions. These systems, however, suffer from various failure cases, and we cannot directly train them end-to-end to fix such failures, as interaction with external knowledge is non-differentiable. To address these deficiencies, we define a ReAct-style LLM agent with the ability to reason and act upon external knowledge. We further refine the agent through a ReST-like method that iteratively trains on previous trajectories, employing growing-batch reinforcement learning with AI feedback for continuous self-improvement and self-distillation. Starting from a prompted large model and after just two iterations of the algorithm, we can produce a fine-tuned small model that achieves comparable performance on challenging compositional question-answering benchmarks with two orders of magnitude fewer parameters.

\end{abstract}

\input{sections/intro}
\input{sections/method}
\input{sections/experiments}

\input{sections/related}
\input{sections/conclusion}



\bibliography{iclr2024_conference}
\bibliographystyle{iclr2024_conference}

\appendix
\input{sections/appendix}

\end{document}

%% file: math_commands.tex
\usepackage{amsmath,amsfonts,bm}

\def\eqref#1{equation~\ref{#1}}

\def\1{\bm{1}}

\DeclareMathAlphabet{\mathsfit}{\encodingdefault}{\sfdefault}{m}{sl}
\SetMathAlphabet{\mathsfit}{bold}{\encodingdefault}{\sfdefault}{bx}{n}



%% file: macros.tex
\newcommand{\renat}[1]{\textcolor{purple}{[Renat: #1]}}
\newcommand{\pranesh}[1]{\textcolor{blue}{[Pranesh: #1]}}
\newcommand{\manzil}[1]{\textcolor{green}{[Manzil: #1]}}
\newcommand{\daliangli}[1]{\textcolor{cyan}{[Daliang: #1]}}
\newcommand{\zkf}[1]{\textcolor{orange}{[Z: #1]}}

%% file: sections/intro.tex
\section{Introduction}

\begin{wrapfigure}{r}{0.55\textwidth} 
    \centering
    \includegraphics[width=0.47\textwidth]{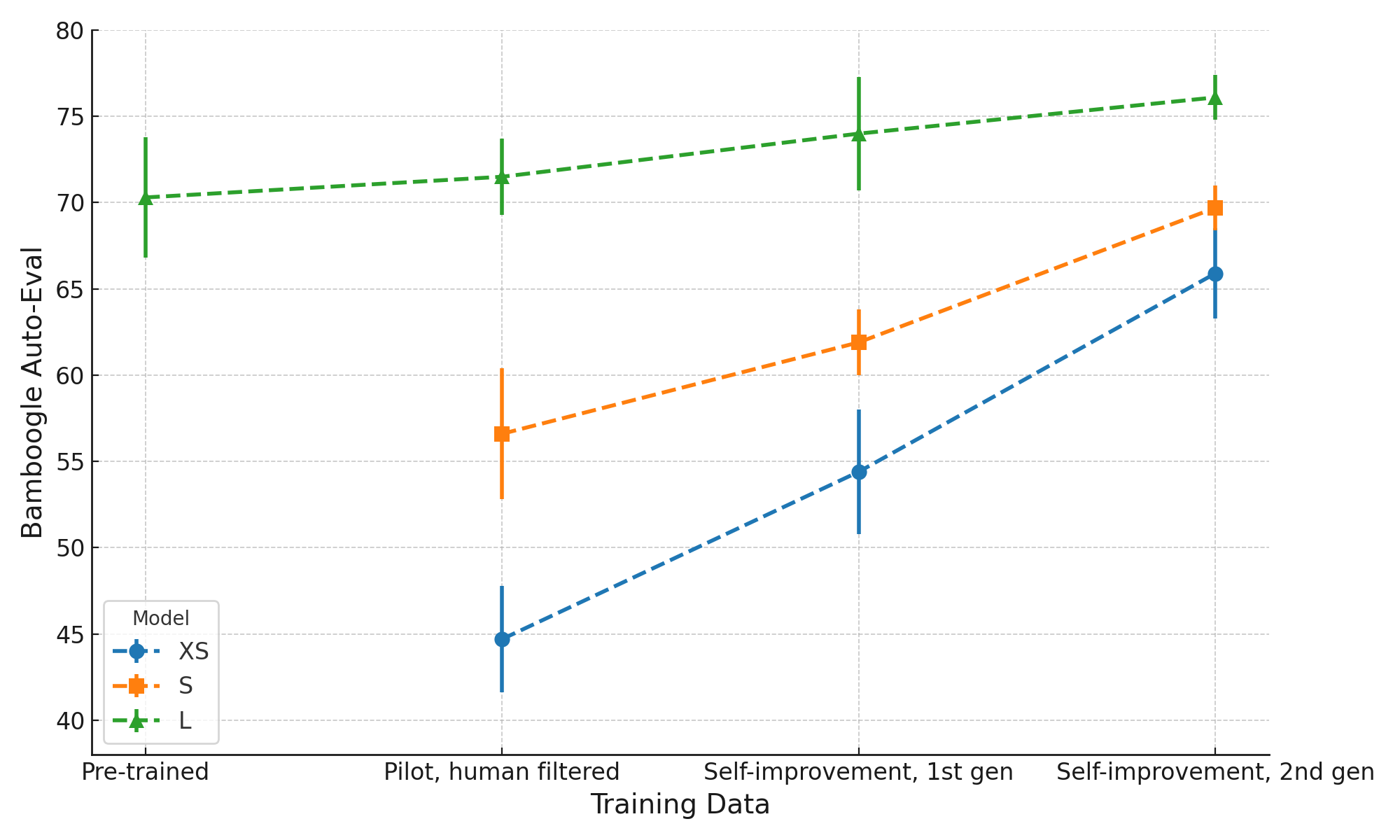}
    \caption{Agent self-improvement and self-distillation. Bamboogle auto-eval, mean accuracy and standard deviation over 10 runs, (\%)}
    \label{fig:llm_progression}
\end{wrapfigure} 

For many simple natural language tasks, like basic question-answering or summarization, we can relatively easily decide whether the final output is good or bad, collect large amounts of such data, and train the language models using these outcomes as feedback. At the same time, for more complex problems, outcome-based systems are often insufficient, and a process supervision approach has recently gained much attention as a more promising alternative (\cite{reppert2023iterated}). There is explosive growth in techniques (\cite{gao2023pal}; \cite{madaan2023selfrefine}), frameworks (\cite{dohan2022language}; \cite{khattab2023dspy}), and libraries (\cite{liu2022llamaindex}, \cite{chase2022langchain}) for defining process-based workflows with LLMs through human-understandable task decompositions. Many such decompositions involve interaction with external tools / APIs / environments, in which case the corresponding multi-step workflow is generally referred to as an LLM agent (\cite{xi2023rise}), a system capable of performing a sequence of actions to achieve a goal.

Let’s consider the task of answering complex, open-ended questions, where the agent needs to use a search API to look up multiple pieces of information before composing a paragraph-length answer. One popular approach for building such agents with LLMs is the ReAct method \citep{yao2022react}, which involves interleaving chain-of-thought reasoning with actions and observations during several thought-action-observation rounds. In this work, we follow the general ReAct format for our Search Agent while designing the corresponding few-shot prompts to produce long-form, explicitly attributable final answers (\textit{cf.} \cite{nakano2021webgpt}).

It is natural to ask next how to deal with failure cases of such an agent and how to improve its performance and robustness. For outcome-based systems, the solution is usually straightforward: we just collect more human-labeled data. However, acquiring such data is much more challenging and expensive for process-based systems: a significantly larger amount of data is needed (\cite{uesato2022solving}; \cite{lightman2023lets}), and it is generally harder for humans to determine an optimal multi-step trajectory.

To address the lack of and difficulty in obtaining multi-step human-labeled data, we focus on improving the quality of the agent with self-critique, AI feedback, and synthetic data generation. Specifically for the latter, we adapt the recently proposed (\cite{gulcehre2023reinforced}) Reinforced Self-Training (\rest) algorithm towards agentic setups. The inner-outer loop flow of \rest remains the same: in the outer loop (“grow”), the dataset is grown by sampling from the latest policy, and in the inner loop (“improve”), the policy is improved on a fixed dataset via ranking or filtering with reward model. In our case, sampling during “grow” means producing a multi-step trajectory to completion, and ranking as part of “improve” is done directly with LLM call rather than with a distilled reward model of human preferences.

We measure the overall performance of the Search Agent by its ability to answer diverse compositional questions that were manually verified to be unanswerable directly by a search engine (Bamboogle dataset \citep{press2023measuring} and a sequel dataset that we constructed ourselves, BamTwoogle). While both datasets are small, they have enough statistical power to capture the effects we are interested in studying. For example, Figure \ref{fig:llm_progression} provides a high-level overview of the agent performance on Bamboogle for different model sizes and showcases the boost with more iterations of \rest (all our synthetic data is produced with the large model, so for smaller models, it is the performance of the distillation).

To summarize, our contributions are the following:
\begin{itemize}
    \item We build a flavor of ReAct agent with self-critique for the task of long-form question answering.
    \item We define a proxy evaluation metric for the agent based on Bamboogle and BamTwoogle datasets, with a strong emphasis on auto-eval.
    \item We demonstrate that the performance of the agent could be effectively improved through Rest-style iterative fine-tuning on its reasoning traces.
    \item Furthermore, we do it purely from stepwise AI feedback without using human-labeled training data.
    \item Finally, we show that the synthetic data produced as part of this iterative process could be used for distilling the agent into one or two orders of magnitude smaller models with performance comparable to the pre-trained teacher agent.
\end{itemize}

%% file: sections/method.tex
\section{Background: Search Agent}
\label{sec:background}

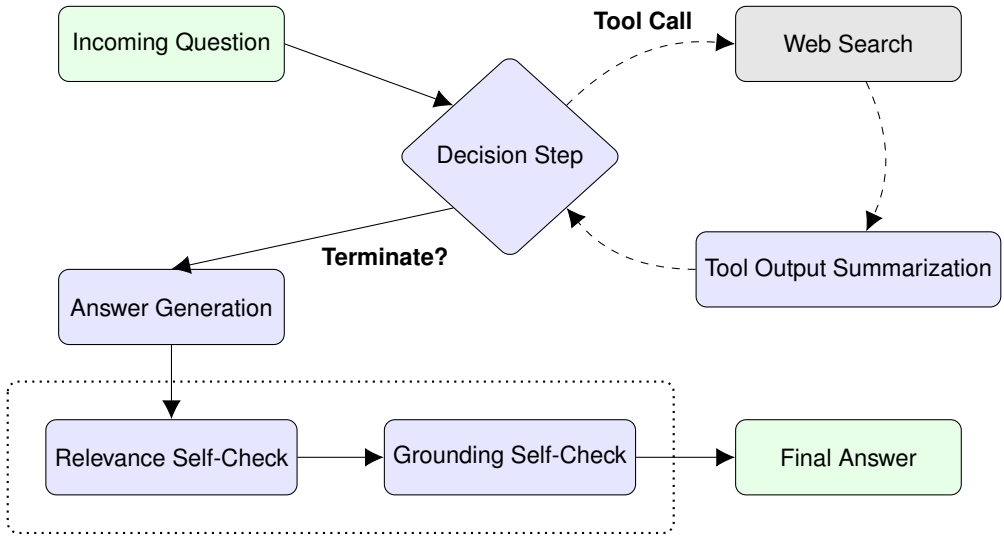
\begin{figure}[ht]
\centering
\begin{tikzpicture}[node distance=2.5cm, on grid, auto,
    every edge/.style={draw,-{Latex[length=2.5mm,width=2.5mm]},align=center, rounded corners},
    every node/.style={font=\sffamily\small},
    state/.style={draw, minimum width=3cm, minimum height=1cm, align=center, rounded corners}]

    \node[state, rectangle, fill=green!10] (incoming) {Incoming Question};
    \node[state, shape=diamond, fill=blue!10, right=of incoming, xshift=2cm, yshift=-1.5cm] (decision) {Decision Step};
    \node[state, rectangle, right=of decision, xshift=2cm, yshift=1.5cm, fill=gray!20] (toolbank) {Web Search};
    \node[state, rectangle, fill=blue!10, below=of toolbank, yshift=-0.5cm] (summary) {Tool Output Summarization};
    \node[state, rectangle, left=of decision, below=of incoming, yshift=-1cm, fill=blue!10] (answer) {Answer Generation};
    \node[state, rectangle, fill=blue!10, below=of answer, yshift=0.5cm] (relevance) {Relevance Self-Check};
    \node[state, rectangle, fill=blue!10, right=of relevance, xshift=2cm] (grounding) {Grounding Self-Check};
    \node[state, rectangle, right=of grounding, xshift=2cm, fill=green!10] (final) {Final Answer};

    \path
        (incoming.east) edge (decision.north west)
        (decision.north east) edge[dashed, bend left=25] node[above=0.2cm] {\textbf{Tool Call}} (toolbank.west)
        (toolbank) edge[dashed, bend left=25] (summary)
        (summary.west) edge[dashed, bend left=25] (decision.south east);
    \path
        (decision.south west) edge node[below right] {\textbf{Terminate?}} (answer.north)
        (answer) edge (relevance)
        (relevance) edge (grounding)
        (grounding) edge (final);
        
    \draw[dotted, thick, rounded corners] ([xshift=-0.5cm,yshift=0.5cm]relevance.north west) rectangle ([xshift=0.5cm,yshift=-0.5cm]grounding.south east);
\end{tikzpicture}
\caption{A state machine of the Search Agent flow. Each blue shape corresponds to a single LLM call and defines a separate type of the reasoning step. 
}
\label{fig:agent-flow}
\end{figure}

This section describes Search Agent, a flavor of ReAct \citep{yao2022react} agent with Reflexion \citep{shinn2023reflexion}. It uses web search as a tool to generate long-form, explicitly attributable answers for diverse knowledge-seeking open-ended questions. The agent's flow proceeds as follows (Figure \ref{fig:agent-flow}):
\begin{enumerate}
    \item Agent receives a question and starts executing a search loop:
      \begin{itemize}
        \item Agent \textbf{decides} whether it needs additional information to answer the question.
        \item If ``yes'', it calls the search tool, \textbf{summarizes} the received snippets, and goes back to the decision step.
        \item If ``no'', it terminates the search loop.
      \end{itemize}
    \item Based on the information collected as part of the search loop, the agent generates the first attempt (draft) of the \textbf{answer}.
    \item It then performs two additional self-revision calls before producing the final answer:
    \begin{itemize}
        \item One to verify that the answer is \textbf{relevant} to the original question,
        \item And another to check that the answer is \textbf{grounded} in the retrieved snippets.
    \end{itemize}
\end{enumerate}

\section{Methods}
\label{sec:methods}

\subsection{Prompting}
We first define the prompted flow for the Search Agent by manually constructing few-shot prompts for each of the five reasoning steps from Figure \ref{fig:agent-flow}. Notably, we choose to format our prompts as Python code (see Appendix for the prompts of different steps, Listings~\ref{lst:decision_step}~-~\ref{lst:self_check_grounding}). 

The decision to use the "code as the prompt" approach is motivated by the following observations:
\begin{itemize}
    \item There is often a need to parse the LLM’s output for integration with other systems and tools, which is much easier to do if the model's input and output are well-structured.
    \item At the same time, code uniquely combines a structured aspect (keywords and syntax) with a natural language aspect (comments and descriptive naming).
    \item Moreover, LLMs are capable of both reading and writing code.
\end{itemize}

To summarize, since code is naturally structured and easy to parse, it could serve as an excellent medium for communication with the model. Accordingly, within the "code as the prompt" paradigm, LLM is expected to understand the code in the input and continue it as valid Python in the output, which turns out to be challenging tasks for the smaller models. Out of the (pre-trained) models we've tried, only the \palmL~\citep{anil2023palm} can perform it consistently well. Hence, this is the model that we use to produce reasoning trajectories from few-shot prompts.

\subsection{Implementation Details}
To run Search Agent, we use \texttt{PaLM 2} ``base'' models of different sizes (\texttt{XS}, \texttt{S} and \texttt{L}), both pre-trained and fine-tuned. We usually produce multiple samples (with $T=0.5$, see Section \ref{sec:auto_eval} for details on temperature selection) for each step and then select the one with the lowest perplexity (4 samples for \texttt{L} and \texttt{S}, 16 for \texttt{XS}) to continue the trajectory forward. As a search tool, we use internal Google Q\&A API that returns top-k snippets, and we typically request top-3 snippets per query from it. We further limit the number of total searches to at most 10 per single agent trajectory (via the REMAINING\_SEARCHES field in the prompts). Search Agent preserves the trajectory state by storing all the actions taken so far in the PAST\_ACTIONS field.

\subsection{Input Data}
We use the following four datasets to provide initial questions for Search Agent trajectories:
\begin{itemize}
    \item \textbf{HotpotQA} \citep{DBLP:journals/corr/abs-1809-09600}, a multi-hop reasoning QA dataset, where the system has to reason with information taken from more than one document to arrive at the answer.
    \item \textbf{Eli5} \citep{DBLP:journals/corr/abs-1907-09190}, a dataset for long-form question answering (LFQA), a task that requires elaborate and in-depth answers to open-ended questions. The dataset was built from the Reddit forum “Explain Like I’m Five” (ELI5), r/explainlikeimfive.
    \item \textbf{Eli5-askH} \citep{blagojevic2022lfqa}, similar to above, but built from the Reddit forum r/askhistorians, a subreddit where users may ask questions or start discussions about history.
    \item \textbf{Eli5-askS} \citep{blagojevic2022lfqa}, as above, but from a subreddit r/askscience (\textit{``ask a science question, get a science answer''}).
\end{itemize}
We randomly selected 500 questions from the training splits of each dataset and ended up with 2000 diverse, challenging questions in total. We don't use any other information from these datasets, like labels (e.g., we don't do any filtering by matching the correct answer from HotpotQA, and we don't use these datasets' validation splits for hyperparameters tuning or performance assessment).

\subsection{Fine-Tuning}
We simply split each completed Search Agent trajectory into the reasoning steps and build a fine-tuning mixture with those steps. We use full fine-tuning for all the experiments. Given that fine-tuning costs increase sharply for larger models, we do as many experiments as possible with \texttt{XS} model.

\subsection{Ranking "Reward" Model}
As mentioned previously, we produce multiple samples for each reasoning step in the agent's trajectory and typically choose the sample that minimizes perplexity to continue the trajectory forward or to build a fine-tuning mixture. At the same time, we might be able to do better than that by utilizing a more sophisticated way of selecting the best sample. To this effect, we employ an instruction-tuned \palmL and prompt it with the model input, multiple sampled outputs, and guidance on how to rank them (the prompt is available in the Appendix, Listing~\ref{lst:reward_model}). We can then use the highest-ranked sample for fine-tuning instead of the default sample chosen based on the perplexity value. 

In this part, our approach differs from \rest, which uses threshold-based filtering with a reward model (RM) trained on human preference data. What we do is closer to RAFT \citep{dong2023raft}, where the reward model ranks sampled responses to select high-scoring subsets for model fine-tuning, and the RM rankings matter much more than absolute scores. Notably, we mainly do the LLM-based selection off-policy, i.e., by updating the current action used for fine-tuning. The on-policy trajectory rollouts use perplexity.

\subsection{Iterative Self-Improvement}
Now we have all the pieces for the self-improvement algorithm: 
\begin{itemize}
    \item Start with a model capable of performing Search Agent task at a certain level, for example, with prompted \palmL model. Collect reasoning trajectories from this model based on our set of 2000 initial questions (essentially the \textbf{``grow''} stage of \rest, with the difference that we keep the set of initial questions fixed).
    \item Convert the trajectories into the fine-tuning mixture. Apply re-ranking with RM during the conversion (this is roughly equivalent to the \textbf{``improve''} stage of \rest, though we only do one iteration of ``improve'').
    \item Fine-tune the new model (of the same size) on this mixture and verify that it's performing better than the original model (we will discuss \textit{how} to do it in the following section). Repeat the process, starting with this new, better model.
\end{itemize}

Finally, we can also train \textit{smaller} models on the fine-tuning data from the different iterations of self-improvement, which will naturally give us a self-distillation algorithm.

\section{Evaluation}
\label{sec:eval}

\subsection{Bamboogle}
Our primary evaluation vehicle is the Bamboogle dataset \citep{press2023measuring}. It is a semi-adversarial dataset of 2-hop questions (\textbf{125} in total) that were selected to be unanswerable by direct Google search, but where both required pieces of evidence could be found in Wikipedia. When the Search Agent's performance on Bamboogle is improving, we can assume that it generally becomes better in using search as a tool.

\begin{wrapfigure}{r}{0.45\textwidth} 
\centering
\begin{tikzpicture}
  \tikzstyle{question}=[rectangle, draw=black, fill=blue!5, text width=6cm, text centered, minimum height=2.5em]
  \tikzstyle{answer}=[rectangle, draw=black, fill=green!5, text width=4cm, text centered, minimum height=2.5em]
  \tikzstyle{modelanswer}=[rectangle, draw=black, fill=red!5, text width=6cm, text centered, minimum height=2.5em]
  \tikzstyle{arrow}=[thick,->,>=stealth]
  
  \node[question] (q) {\textbf{Question}: What is the maximum airspeed (in km/h) of the third fastest bird?};
  \node[modelanswer, below=of q] (ma) {\textbf{Model Answer}: The golden eagle is the third fastest bird. According to [link\_id=4, 5, 6], its maximum airspeed is 200mph, which is 320kph.};
  \node[answer, below=of ma] (a) {\textbf{Ground truth}: 320 km/h};
  
  \draw[arrow] (q) -- (ma);
\end{tikzpicture}
\caption{A Bamboogle question with an example of the long form answer by the model}
\label{fig:bamboogle-example}
\end{wrapfigure}
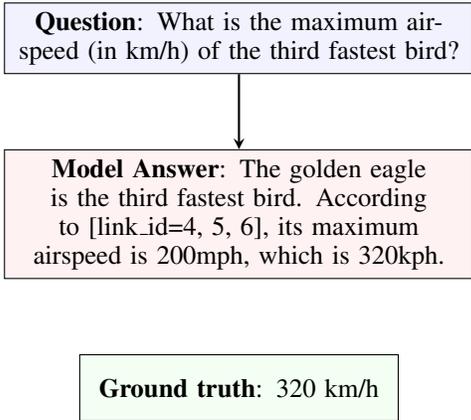

Given the open-ended nature of the answers generated by the Search Agent (Figure~\ref{fig:bamboogle-example}), we can't reliably use the exact match as a metric. Instead, we either manually judge correctness or run auto-eval with a separate call to the \palmL ``base'' model.

\subsection{Auto-Eval}
\label{sec:auto_eval}

While strict human evaluations are preferable, they are time-consuming. They also don't scale well (doing one human eval is much easier than doing five), which leads to a high variance of such evals in our case: the agent's trajectories are stochastic (as a reminder, we use non-zero temperature when sampling reasoning steps), but we can't easily reduce the variance by increasing the number of repetitions per question with human evals.

We solve both of these problems by introducing LLM-based auto-eval  (the full auto-eval prompt is available in the Appendix, Listing \ref{lst:auto_eval}) and by verifying its alignment with human ratings for the Bamboogle trajectories on which we previously conducted human evals. We compared the auto-eval with human evaluations on a diverse set of agents and found it highly correlated with human evaluation scores. Specifically, the Pearson correlation is 0.98 with $p = 6.6 \times 10^{-8}$ and Spearman correlation is 0.83 with $p = 0.0015$. Given that it is much cheaper to run auto-eval, we can now use a large number of repetitions to reduce variance. We typically aggregate auto-eval over ten repetitions (i.e., by producing ten different trajectories for each Bamboogle question).

\begin{wrapfigure}{r}{0.55\textwidth}
    \centering
    \includegraphics[width=0.47\textwidth]{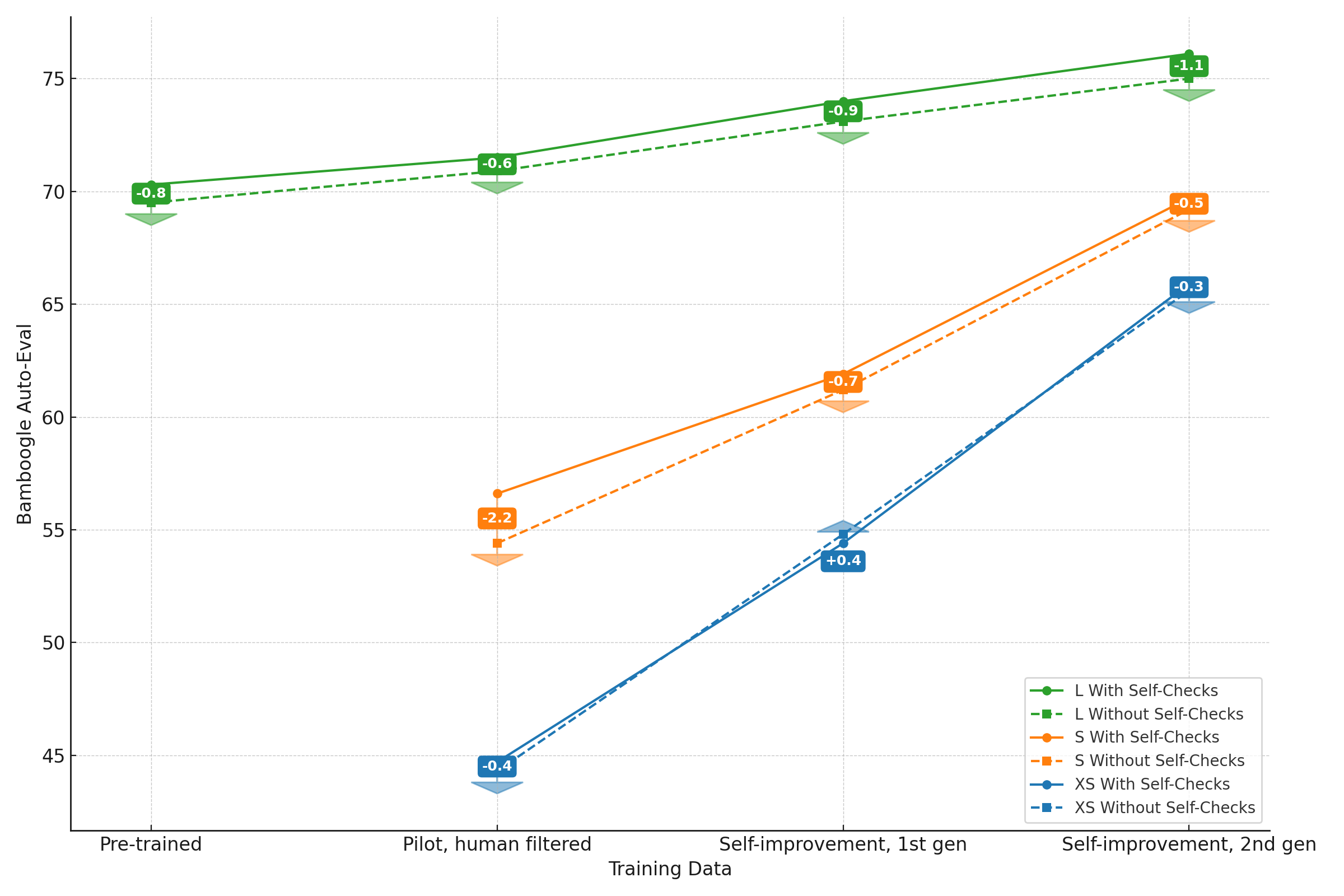}
    \caption{Agent self-improvement and self-distillation, with and without self-critique steps. Bamboogle auto-eval, mean accuracy and standard deviation over ten runs, (\%)}
    \label{fig:llm_progression_self_checks}
\end{wrapfigure}

First and foremost, we use Bamboogle auto-eval to estimate the final model performance but also to answer various questions that one would typically use a validation set for:
\begin{itemize}
    \item What is the optimal sampling temperature for the agent? ($T=0.5$)
    \item Which checkpoints should we choose for different model sizes? (step 9K for \texttt{XS}, 5K for \texttt{S}, 3.5K for \texttt{L})
    \item Should we proceed with another iteration of self-improvement?
    \item What is the performance impact of using several trajectories per question on the fine-tuned model?
    \item Should we use self-checks? Are they helping or hurting the results? (helping slightly, see Figure~\ref{fig:llm_progression_self_checks})
\end{itemize}

Importantly, we never use Bamboogle as a training set, as we neither tune our prompts on Bamboogle nor use questions from it to generate fine-tuning trajectories. 

\subsection{BamTwoogle}

Given the small size of Bamboogle and our use of it as an analog of the validation set for Search Agent, there is a danger of overfitting. To guard against such a possibility, we introduce a new dataset, BamTwoogle, to serve as a test set. We use BamTwoogle exclusively to measure the final performance of the models.

The BamTwoogle dataset was written to be a complementary, slightly more challenging sequel to Bamboogle. It also addresses some of the shortcomings of Bamboogle we discovered while performing human evals. Specifically, we ensured that all the questions required 2+ steps to answer. Due to changes in the search algorithm, this is no longer the case with all Bamboogle questions, and it is possible to solve some of them with a single search.

Like Bamboogle, BamTwoogle is a small (\textbf{100} questions in total), handcrafted collection of information-seeking questions. The topics and question formats vary, but in general, BamTwoogle adheres to the following guidelines

\textbf{Questions}\nopagebreak
\begin{itemize}
    \item The majority of questions require two searches or reasoning steps (like Bamboogle), but some of them need 3 or 4
    \item Must have been manually checked to ensure the answer doesn’t appear on the first page of Google search results
\end{itemize}

\textbf{Expected answers}\nopagebreak
\begin{itemize}
    \item Should not be ambiguous
    \item Should not be prone to change over time, either due to the phrasing of the question or to the nature of the answer
    \item Should account for multiple versions of proper names, etc., where appropriate
    \item Should prefer Wikipedia as the source of truth for facts (preference given to topics/articles not flagged for incompleteness, lack of sources, etc.)
\end{itemize}


%% file: sections/experiments.tex
\section{Experiments}
\label{sec:exp}

\renewcommand{\arraystretch}{1.2} 

\subsection{Pilot}

\begin{wraptable}{r}{0.6\textwidth}
\captionsetup{width=0.55\textwidth}
\caption{Agent self-improvement and self-distillation, Bamboogle auto-eval, mean accuracy and standard deviation over 10 runs, (\%)}
\label{tab:overall}
\small
\setlength{\tabcolsep}{4pt}
\begin{tabular}{l|ccc}
\toprule
 Training Data & \texttt{XS} & \texttt{S} & \texttt{L} \\
\midrule
Pre-trained & N/A & N/A & 70.3\textsuperscript{$\pm 3.5$} \\
\midrule
Pilot, human filtered & 44.7\textsuperscript{$\pm 3.1$} & 56.6\textsuperscript{$\pm 3.8$} & 71.5\textsuperscript{$\pm 2.2$} \\
Self-improvement, 1st gen & 54.4\textsuperscript{$\pm 3.6$} & 61.9\textsuperscript{$\pm 1.9$} & 74.0\textsuperscript{$\pm 3.3$} \\
Self-improvement, 2nd gen & 65.9\textsuperscript{$\pm 2.6$} & 69.7\textsuperscript{$\pm 1.3$} & 76.1\textsuperscript{$\pm 1.3$} \\
\bottomrule
\end{tabular}
\end{wraptable}

Alongside the main self-improvement setup, described in Section~\ref{sec:methods}, we also generate a simpler ``pilot'' data with 500 trajectories, where the initial questions are selected from HotpotQA and Eli5 datasets only (i.e., smaller and without Eli5-askH or Eli5-askS). We use the default (based on a min perplexity, no RM re-ranking) best actions for building the pilot's fine-tuning mixture. Furthermore, we manually review the fine-tuning data and filter out about $30\%$ of the examples that are "bad" in some way: an unhelpful query, empty thoughts, summary missing important information, etc. This pilot data serves as a fine-tuning baseline.

\subsection{Self-improvement and self-distillation}

The main results are presented in Figure~\ref{fig:llm_progression} and Table~\ref{tab:overall}. As discussed previously, each cell in the table shows a mean (with the corresponding std) of auto-eval over 10 Bamboogle runs for the particular model.

\begin{wraptable}{r}{0.7\textwidth}
\captionsetup{width=0.65\textwidth}
\caption{Human evals, Bamboogle and BamTwoogle, acc (\%)}
\label{tab:human-evals}
\small
\setlength{\tabcolsep}{4pt}
\begin{tabular}{l|c|ccc}
\toprule
 & \multicolumn{1}{c}{Pre-trained (\texttt{L})} &  \multicolumn{1}{c}{2nd gen (\texttt{XS})} & \multicolumn{1}{c}{2nd gen (\texttt{S})} & \multicolumn{1}{c}{2nd gen (\texttt{L})} \\
\midrule
Bamboogle & 68.8 & 67.2 & 68.0 & 74.4 \\
\midrule
BamTwoogle & 68.0 & 63.0 & 63.0 & 74.0 \\
\bottomrule
\end{tabular}
\end{wraptable}

We start with a pre-trained (prompted) \palmL model and use it to generate both 500 pilot trajectories and (independently) 2000 trajectories for 1st iteration ("1st gen") of self-improvement. We then fine-tune \palmL, \palmS, and \palmXS models on the resulting mixtures.

Next, we use the \palmL model fine-tuned on 1st gen data to generate trajectories for the 2nd iteration ("2nd gen") of self-improvement. We build 2nd gen data with 8000 trajectories (using the same 2000 initial questions, each repeated four times; analogous to a "grow" stage in \rest). As previously, we fine-tuned each of the three models on the new 2nd gen mixture.

As a final verification, we also do human evals over a single Bamboogle and BamTwoogle run for each of the 2nd gen models (Table~\ref{tab:human-evals}), as well as the original pre-trained \palmL model for comparison.

\subsection{Ablations}

\paragraph{What is the effect of human filtering?} Surprisingly, we have found that fine-tuning on filtered data results in a small performance drop (2.5\%) versus unfiltered pilot mixture (Table~\ref{tab:data}, pilot columns). We hypothesize that this happens due to a combination of 2 factors:
\begin{itemize}
    \item The reduced size of the filtered mixture makes it harder for the model to learn the proper format of the prompts,
    \item Our filtering only affects the immediate "bad" example, not the whole trajectory; the "bad" step would often be preserved in the other fine-tuning examples as part of the PAST\_ACTIONS field.
\end{itemize}

\paragraph{Should we use multiple trajectories per question?} Turns out, it helps to use two trajectories per question instead of 1 (2.2\% gain) in the fine-tuning mixture, but more than that doesn't improve performance significantly (Table~\ref{tab:data}, 2nd gen columns).

\begin{table}[h]
\centering
\caption{Impact of the quality and size of the training data for fine-tuned \palmXS models}
\label{tab:data}
\small
\setlength{\tabcolsep}{4pt}
\begin{tabular}{l | p{1.7cm} p{1.4cm} | l | lll}
\toprule
 & \parbox{1.7cm}{Pilot, human \\ filtered} & \parbox{1.4cm}{Pilot,\\ unfiltered} & 1st gen & 2nd gen (1x) & 2nd gen (2x) & 2nd gen (4x) \\
\midrule
Total trajectories & 500 & 500 & 2000 & 2000x1 & 2000x2 & 2000x4 \\
Training examples & 3015 & 4518 & 17970 & 18007 & 36238 & 72424 \\
\midrule
Bamboogle auto-eval & 44.7\textsuperscript{$\pm 3.1$} & 47.2\textsuperscript{$\pm 3.1$} & 54.4\textsuperscript{$\pm 3.6$} & 63.4\textsuperscript{$\pm 1.7$} & 65.6\textsuperscript{$\pm 1.8$} & 65.9\textsuperscript{$\pm 2.6$} \\
\bottomrule
\end{tabular}
\end{table}

\paragraph{More data vs better data.} From the same Table~\ref{tab:data}, we can conclude that the quality of the data (e.g., 9\% gain, when going from 1st gen to 2nd gen (1x) while keeping the size of the data roughly the same) matters more than its quantity. Notably, better data also reduces the variance of evaluation trajectories.

\paragraph{Effect of self-critique.} The multi-step setup of the agent allows us to easily measure the effect of the self-critique steps on the overall agent performance. To do this, we simply take all the Bamboogle trajectories used for Table~\ref{tab:overall} and run auto-eval on them, but over the "Answer Generation" step, rather than over "Final Answer", as we normally do (Figure~\ref{fig:agent-flow}). As seen from Figure~\ref{fig:llm_progression_self_checks}, the self-critique results in a small but measurable positive boost (on the order of 0.5-1.0\% for most models). The detailed numbers are provided in Table~\ref{tab:overall-no-critique} in the Appendix.

\section{Discussion}
\label{sec:discuss}

\paragraph{Process Supervision.} It's important to emphasize again that we don't use labels from the training data as a signal during trajectory collection. This is possible due to combining a process-based approach (i.e., defining agent as a state machine) with high-temperature exploration, AI feedback (zero-shot "reward" model used for actions re-ranking), and state-wise fine-tuning over completed trajectories. In other words, in this setup, the model can learn something useful even from the states that eventually lead to the wrong final answer. More importantly, it learns to handle open-ended questions that don't have a single well-defined correct answer in the first place, akin to many questions from Eli5.

\paragraph{Auto-Eval.} Some of the properties listed above, like non-greedy sampling and long-form answer generations in particular, bring with them additional challenges in terms of agent evaluation. We both need to measure the quality of the long-form final answer for a specific trajectory and account for stochasticity between different agent trajectories for the same input. This, on the one hand, elevates the value of having robust auto-eval aligned with human raters and, on the other, increases the computational costs significantly due to the need to run agent trajectories multiple times, as well as the use of \palmL model for auto-eval.

\paragraph{Self-Critique.} Despite those computational costs, multiple benefits come from having reliable auto-eval in terms of measuring the impact of various agent's hyperparameters. As an example, let's consider the self-critique steps. With the aid of auto-eval, we were able to establish that they have a small but positive effect on the overall performance of our multi-step reasoning setup. This contrasts the recent observations \citep{huang2023large} that adding self-critique hurts the performance in an outcome-based CoT setup. Simultaneously, we can also notice that a positive effect of self-critique depends on the model size (larger for larger models) but does not seem to be affected by the self-improvement process.

\paragraph{Limitations and Future Directions.} While this work lays a foundation for reasoning agents' self-improvement, it has numerous limitations due to computational and time constraints: manually constructed prompts, small evals, a limited set of models, and only a single tool, to name just a few.

Future work could explore if the same self-improvement algorithm applies to multiple tool settings and, especially, if the ability to handle unseen tools could be improved in such a way. If the latter is more similar to self-critique and doesn't improve under \rest-like iterative training, what changes are required to enable self-improvement for both? 

Another open question is a saturation point. How many additional iterations of self-improvement can we undertake past the 2nd one that still results in non-trivial gains? What does the saturation look like for smaller models? Will they all eventually converge to the same performance, or will the smaller models always be capped by the performance of the initial prompted large model?

%% file: sections/related.tex
\section{Related Work}
\label{sec:related}

Following WebGPT \citep{nakano2021webgpt}, we are tackling the task of long-form question answering \citep{krishna2021hurdles}, in which the language agent uses web search as a tool to generate final answers with explicit references for the retrieved passages. While WebGPT is focused on imitation learning and RL from a large number of human demonstrations, our work aims to minimize human involvement. The only labeled demonstrations we use as part of the training are few-shot exemplars in the prompts for the agent’s reasoning steps (see Appendix, \ref{sec:steps}).

Setting up language agents with manually designed few-shot prompts is the most common practice (\cite{press2023measuring}; \cite{yao2022react}; \cite{shinn2023reflexion}), but there are some exceptions. For example, DSP \citep{khattab2023demonstratesearchpredict} tunes the few-shot demonstrations for the prompts automatically, leveraging some amount of labeled training examples for optimization purposes, and can further fine-tune specific components of the agent.

Unlike prompting, agent’s fine-tuning is done less often (\cite{nakano2021webgpt}; \cite{yao2022react}; \cite{chen2023fireact}). The closest to our fine-tuning setup is probably FireAct \citep{chen2023fireact}, with the main difference being that we don’t use human labels for training or data filtering. Instead, we are building synthetic data with self-improvement from AI feedback.

Some relevant papers for self-improvement include STAR \citep{zelikman2022star}, \rest \citep{gulcehre2023reinforced}, \rest{\textsuperscript{EM}} \citep{singh2023human}, and RAFT \citep{dong2023raft}. Unlike STAR and \rest{\textsuperscript{EM}}, we don’t use the correctness of the answer as a signal. And, unlike \rest and RAFT, we don’t have the proper reward model trained on human preferences. Moreover, all 4 of these papers target outcome-based systems, while we focus on a process-based one.

%% file: sections/conclusion.tex
\section{Conclusion}
\label{sec:conclusion}

This work demonstrates that the \rest-like approach with AI feedback could be effectively applied to a multi-step reasoning LLM agent. We show that it is a relatively simple and efficient way to iteratively build high-quality synthetic data for agent self-improvement. Moreover, this increasingly higher quality data could simultaneously be used for distilling a multi-step agent into several magnitudes smaller models while preserving most of the performance from the large teacher model.

%% file: sections/appendix.tex
\section{Appendix}
\label{sec:appendix}

\subsection{Prompts}

\paragraph{Reasoning steps.}
\label{sec:steps}

All our reasoning prompts are n-shot (with $n > 1$), though we only show fragments, limited to the first exemplar:
\begin{itemize}
    \item The \textbf{decision} step prompt is shown in Listing~\ref{lst:decision_step} (the full prompt is 9-shot).
    \item The \textbf{summarization} prompt is presented in Listing~\ref{lst:summarization_step} (the full prompt is 6-shot).
    \item The \textbf{answer generation} prompt is 5-shot, Listing~\ref{lst:answer_generation_step} shows a 1-shot fragment.
    \item The prompt for \textbf{relevance self-check} is presented in Listing~\ref{lst:self_check_relevance} (the full prompt is 6-shot).
    \item Finally, the prompt for \textbf{grounding self-check} is shown in Listing~\ref{lst:self_check_grounding} (the full prompt is 5-shot).
    \item Both self-checks use the common prefix from Listing~\ref{lst:self_check_prefix}.
\end{itemize}

\begin{figure}[h]
\centering
\begin{lstlisting}[style=mypy, caption=Decision step prompt (fragment), label=lst:decision_step]
"""Implement an agent capable of answering complex queries by potentially search multiple times.
"""
import dataclasses

class Action:
  """Base class for different actions."""
  ...

@dataclasses.dataclass
class ActionWrapper:
  """Encapsulates the reasoning as well as the selected action.

  Attributes:
    thoughts: Record your thoughts on why we should do this action.
    action: The actually selected action.
  """
  thoughts: str
  action: Action

@dataclasses.dataclass
class Search(Action):
  """The Google search command.

  Attributes:
    query: The query to be sent to Google.
  """
  query: str

@dataclasses.dataclass
class Terminate(Action):
  """Command to terminate the search sequence."""
  ...

#########################
# Example 1:
#########################

ORIGINAL_QUESTION: str = 'I want to buy the Ikea Klippan loveseat, but I\'m not sure whether it can fit into my 2019 Honda Odyssey. Could you check whether I need to fold down the seat?'
PAST_ACTIONS: List[Action] = [
Search(query='ikea klippan loveseat dimension',
thoughts="To answer the ORIGINAL_QUESTION, we need to find the size of the Ikea Klippan loveseat and the cargo size of the 2019 Honda Odyssey, and then compare them. Let's start by finding the size of the Ikea Klippan loveseat.",
),
SelectLink(selected_links=[
  ResultItem(link_id=1, link_text='KLIPPAN Loveseat, Vissle gray - IKEA',
             snippet='<lh> Measurements </lh> <ul> <li> Width: 70 7/8 &quot; </li> <li> Depth: 34 5/8 &quot; </li> <li> Height: 26 &quot; </li> <li> Height under furniture: 4 3/8 &quot; </li> <li> Seat depth: 21 1/4 &quot; </li> <li> Seat height: 16 7/8 &quot; </li> </ul>'),],
grounded_summarization='According to [link_id=1] the dimension of the ikea klippan loveseat is Width: 70 7/8"; Depth: 34 5/8"; Height: 26".',
thoughts="We still need to find the cargo size of the 2019 Honda Odyssey to answer the ORIGINAL_QUESTION.",
),
]
REMAINING_SEARCHES: int = 4
ORIGINAL_QUESTION: str = 'I want to buy the Ikea Klippan loveseat, but I\'m not sure whether it can fit into my 2019 Honda Odyssey. Could you check whether I need to fold down the seat?'

ACTION_SELECTED = ActionWrapper(thoughts="The past result gives us the dimension of the love seat. We indeed need to find the cargo size of the 2019 Honda Odyssey.", action=Search(query='2019 Honda Odyssey cargo size'))  # [END]

#########################
# Example 2:
#########################
\end{lstlisting}
\end{figure}

\begin{figure}[h]
\centering
\begin{lstlisting}[style=mypy, caption=Summarization prompt (fragment), label=lst:summarization_step]
"""Implementing step 2 of Search and Answer procedure: Search result filtering and summarization.
"""

#########################
# Example 1:
#########################

ORIGINAL_QUESTION: str = "What's the efficiency of the solar panels on the garden solar lights?"
PAST_ACTIONS: List[Action] = [
Search(query='garden solar light panel efficiency',
thoughts="Let's just rephrase the search query a bit. The intention of the original question is pretty clear.",
),
]
CURRENT_SEARCH_RESULTS = SearchResult(links=[
  ResultItem(link_id=17, link_text='Use Solar Outdoor Lighting for Energy Efficiency - Sepco Solar', 
             snippet='Depending on the type of lighting you install and the wattage of the fixtures, this can eat up a budget quickly. Instead, going with a solar outdoor lighting system to provide the required light can greatly reduce the project costs, help improve the environment, and keep electrical costs in the bank where it belongs.'),
  ResultItem(link_id=18, link_text='How Do Solar Garden Lights Work? - The OutdoorLights Store', 
             snippet='The solar panel converts solar energy into electricity<br><br> The energy absorbed by the photovoltaic cells in the panel creates electrical charges. These charges move in response to an electrical field in the solar panel&#39;s cells, which ultimately causes electricity to flow.'),
  ResultItem(link_id=19, link_text='10 things you should know about the garden solar lights', 
             snippet='The solar panels are generally made from two types of cells, monocrystalline or polycrystalline. Both of them are low cost, however, different panel technology can lead to different efficiency under different conditions.'),
])
ORIGINAL_QUESTION: str = "What's the efficiency of the solar panels on the garden solar lights?"

# [link_id=17] is an article about garden solar light, which doesn't have anything specific about the solar panels. Not selected.
# [link_id=18] is an article on how to best install those solar lights. Not relevant to the solar panels used. Not selected.
ACTION_SELECTED: LinkSelection = LinkSelection(grounded_summarization='According to [link_id=19], there are two types of solar cells used for garden solar lights, the monocrystalline or polycrystalline panels.', thoughts="It seems like non of the links shows the efficiency number. However, link 19 provides some hints on what to search next. We can look into the efficiency number of those two types of panels.", selected_link_ids=[19])  # [END]

#########################
# Example 2:
#########################
\end{lstlisting}
\end{figure}

\begin{figure}[h]
\centering
\begin{lstlisting}[style=mypy, caption=Answer generation prompt (fragment), label=lst:answer_generation_step]
"""Implementing step 3 of the multi-round search agent: Answer generation based on all the collected search results.
"""

#########################
# Example 1:
#########################

ORIGINAL_QUESTION: str = "What's the white powdery residual often found on a humidifier?"
PAST_ACTIONS: List[Action] = [
Search(query='what is humidifier',
thoughts="Let's solve this step by step. The person is asking about the residual found in humidifier. Let's first define what does humidifier do.",
),
SelectLink(selected_links=[
  ResultItem(link_id=1, link_text='Humidifiers: What They Do for Health, Uses, Types, and More',
             snippet='What is a humidifier? Humidifiers are devices that add moisture to the air to prevent dryness that can cause irritation in many parts of the body. Humidifiers can be particularly effective for treating dryness of the skin, nose, throat, and lips. They can also ease some symptoms caused by the flu or common cold.'),
  ResultItem(link_id=3, link_text='5 humidifier uses: Benefits and risks - Medical News Today',
             snippet='We do the research so you can find trusted products for your health and wellness. Read more about our vetting process. Humidifiers add moisture to the air, which can benefit people with respiratory symptoms or dry skin. There are several ways to use humidifiers in the home or office, but there are also some risks.'),],
grounded_summarization='According to [link_id=1] and [link_id=3], humidifier is a device that adds moisture to the air to prevent dryness that can cause irritation in many parts of the body.'),
thoughts="Now that we know what humidifier is, we can look into the white powdery residual.",
),
Search(query='white powdery residual in humidifier',
thoughts="Let's now look into the white powdery residual.",
),
SelectLink(selected_links=[
  ResultItem(link_id=10, link_text='What is white dust and should I be concerned about it? - HVAC.com',
             snippet='White dust is usually caused by mineral content in the water that goes into a humidifier. When the mist lands on furniture or other surfaces and dries, it can sometimes leave behind dried mineral deposits, or "white dust."'),
             link_text='What is white dust and should I be concerned about it? - HVAC.com', snippet='White dust is usually caused by mineral content in the water that goes into a humidifier. When the mist lands on furniture or other surfaces and dries, it can sometimes leave behind dried mineral deposits, or "white dust."')
  ResultItem(link_id=11, link_text='Why is White Dust Coming Out of My Humidifier?',
             snippet='The white dust that you are finding on surfaces around your home is likely dried-up minerals from water vapor created by the humidifier. These minerals are natural in our home water supply. If you&#39;ve ever heard of the term "hard water," it&#39;s referring to water with high levels of calcium and magnesium.'),],
grounded_summarization='According to [link_id=10], and [link_id=11], the white powdery residual is caused by the mineral content in the water.'),
thoughts="We have all the information to answer the question.",
),
Terminate(thoughts="We have all the information to answer the question."),
]
ORIGINAL_QUESTION: str = "What's the white powdery residual often found on a humidifier?"

ACTION_SELECTED: Answer = Answer(thoughts="The first search ('what is humidifier') is probably not what the questioner cares, so let's omit that in the answer. We can focus more on the question itself.", answer="The white powdery residue in the humidifier is usually caused by mineral content in the water, according to [link_id=10]. For example, water may contain high levels of calcium and magnesium [link_id=11].")  # [END]

#########################
# Example 2:
#########################
\end{lstlisting}
\end{figure}

\begin{figure}[h]
\centering
\begin{lstlisting}[style=mypy, caption=Self-check prompt (prefix), label=lst:self_check_prefix]
"""Check whether the ANSWER addresses the ORIGINAL_QUESTION and whether the ANSWER is based on SelectLinks in PAST_ACTIONS."""
from dataclasses import dataclass
from typing import List, Tuple

class Action:
  """Base class for different actions."""

class Check_Answer(Action):
  """Check whether the ANSWER addresses the ORIGINAL_QUESTION."""

  def __init__(self, passed: bool) -> None:
    self.passed = passed

class Revise_Answer(Action):
  """Revise the answer if it did not pass the check, based on information from SelectLinks in PAST_ACTIONS."""

  def __init__(self, revised_answer: str) -> None:
    self.revised_answer = revised_answer

  ...

@dataclasses.dataclass
class ResultItem:
  """Single search result, with link id, link title and snippet.

  Attributes:
    link_id: A unique integer id of this link.
    link_text: The title of the link.
    snippet: The snippet from the page that's relevant to the query
  """

  link_id: int
  link_text: str
  snippet: str

\end{lstlisting}
\end{figure}

\begin{figure}[h]
\centering
\begin{lstlisting}[style=mypy, caption=Relevance self-check (fragment), label=lst:self_check_relevance]
#########################
# Example 1:
#########################

ORIGINAL_QUESTION: str = 'How to exclude a website from Google search'
PAST_ACTIONS: List[Action] = [
Search(query='exclude site in google search',
thoughts="We simply the ORIGINAL_QUESTION to a good search query.",
),
SelectLink(selected_links=[
  ResultItem(link_id=2, link_text='How to Exclude a Website from Google Search Results - Keyword',
             snippet="You can exclude a domain or even a specific page from appearing on Google search results. Here's how ..."),
  ResultItem(link_id=3, link_text='How do I exclude/block specific web sites from search results?'),
             snippet="In the absence of a browser extension, the simplest way is to add -site:website_name after your search terms. ... I can't get rid of books.google.com results. I' ..."),
  ResultItem(link_id=6, link_text="Q&A: Can you exclude a specific website from Google's search results?"),
             snippet='Jul 31, 2022 . All you have to do is simply append -site:pinterest.com to the end of your search query. For example, if you were searching for the term "paper ...'),
  ],
grounded_summarization='[link_id=3] mentions that we can add -site:website_name to the google search query to exclude a website. [link_id=6] provides an example.'),
thoughts="We found enough info for a good answer to the ORIGINAL_QUESTION.",
),
Terminate(thoughts="This is a good answer to the question.")
Answer(thoughts="We summarize the relevant info found in PAST_ACTIONS.", answer="To exclude a website from Google search result, you can add -site:website_name after your search terms according to [link_id=3]. For example, if you want to exclude pinterest.com, you can simply add -site:pinterest.com according to [link_id=6].")
]
ORIGINAL_QUESTION: str = 'How to exclude a website from Google search'
ANSWER: str = 'To exclude a website from Google search result, you can add -site:website_name after your search terms according to [link_id=3]. For example, if you want to exclude pinterest.com, you can simply add -site:pinterest.com according to [link_id=6].'

# The ANSWER is directly addressing the ORIGINAL_QUESTION.
# The ANSWER refers to [link_id=3] which correctly supports the core statement of adding -site:website_name. The example from the answer is indeed from [link_id=6].
ACTION_SELECTED: Command = Check_Answer(passed=True)  # [END]

#########################
# Example 2:
#########################
\end{lstlisting}
\end{figure}

\begin{figure}[h]
\centering
\begin{lstlisting}[style=mypy, caption=Grounding self-check (fragment), label=lst:self_check_grounding]
#########################
# Example 1:
#########################

ORIGINAL_QUESTION: str = "What did Jim Betts' competitor in the 1980 election achieve in 1962?"
PAST_ACTIONS: List[Action] = [
Search(query='who was Jim Betts competitor in 1980',
thoughts="This question is asking about the achievement of Jim Betts' competitor in 1962. Let's first search for who this competitor is.",
),
SelectLink(selected_links=[
  ResultItem(link_id=1, link_text='Jim Betts (politician) - Wikipedia',
             snippet='Jim Betts is a former member of the Ohio House of Representatives. He ran against incumbent U.S. Senator John Glenn in 1980. He lost and ran in 1982 for lieutenant governor.'),],
grounded_summarization="According to [link_id=1], Jim Betts' competitor in 1980 was John Glenn."),
thoughts="Now that we know who the competitor is, we can search for his achievements in 1962.",
),
Search(query='John Glenn achievements 1982',
thoughts="Now that we know who the competitor is, we can search for his achievements in 1982.",
),
SelectLink(selected_links=[
],
grounded_summarization='Nothing is selected.'),
thoughts="The query 'John Glenn achievements 1982' is not very good. It should be 'John Glenn achievements 1962'.",
),
Search(query='John Glenn achievements 1962',
thoughts="John Glenn is the person we are looking for. Let's change the date to 1962.",
),
SelectLink(selected_links=[
  ResultItem(link_id=19, link_text='John Glenn | Biography, Accomplishments, & Facts | Britannica',
             snippet="Glenn, Jr., (born July 18, 1921, Cambridge, Ohio, U.S.-died December 8, 2016, Columbus, Ohio), the first U.S. astronaut to orbit Earth, completing three orbits in 1962. (Soviet cosmonaut Yuri Gagarin, the first person in space, had made a single orbit of Earth in 1961.) Glenn joined the U.S. Naval Reserve in 1942."),],
grounded_summarization="According to [link_id=19], John Glenn's achievement in 1962 was to become the first American to orbit the Earth."),
thoughts="We have all the information to answer the question.",
),
Terminate(thoughts="We have all the information we need to answer the ORIGINAL_QUESTION. Let's stop searching and answer the question.")
Answer(thoughts="We have the answer to the question.", answer="Jim Betts\' competitor in the 1980 election was John Glenn. And John Glenn\'s achievement in 1962 was to become the first American to orbit the Earth.")
]
ORIGINAL_QUESTION: str = "What did Jim Betts' competitor in the 1980 election achieve in 1962?"
ANSWER: str = 'Jim Betts\' competitor in the 1980 election was John Glenn [link_id=1]. And John Glenn\'s achievement in 1962 was to become the first American to orbit the Earth [link_id=19].'

# ANSWER directly addresses the ORIGINAL_QUESTION.
# [link_id=1] indeed says that Jim Betts\' run against John Glenn in 1980, consistent with ANSWER.
# [link_id=19] indeed says John Glenn is the first US astronaut to orbit the earth, consistent with ANSWER.
ACTION_SELECTED: Command = Check_Answer(passed=True)  # [END]

#########################
# Example 2:
#########################
\end{lstlisting}
\end{figure}

\paragraph{Auto-eval.} The full 5-shot prompt for auto-eval is presented in Listing~\ref{lst:auto_eval}.

\begin{figure}[h]
\centering
\begin{lstlisting}[style=mypy, caption=Auto-eval (full prompt), label=lst:auto_eval]
"""Check whether the ANSWER implies the REF_ANSWER to the question."""

def Check_Answer(ORIGINAL_QUESTION, ANSWER, REF_ANSWER):
  """Check whether the ANSWER implies the REF_ANSWER."""
  # todo

#########################
# Example 1:
#########################

ORIGINAL_QUESTION: str = 'Are McIntyre Mines and Copperfields Mine in the same country?'
ANSWER: str = 'According to [link_id=1], Copperfields Mine is in Ontario, Canada. According to [link_id=4], McIntyre Mines is also in Ontario, Canada. So yes, they are in the same country.'
REF_ANSWER: str = 'yes'

# the ANSWER implies the answer to the original question is yes, this is consistent with the REF_ANSWER.
Check_Answer(ORIGINAL_QUESTION, ANSWER, REF_ANSWER) = True  # [END]

#########################
# Example 2:
#########################

ORIGINAL_QUESTION: str = 'Which Director, James Richard Kelly or Paul L. Stein has a list of 67 films to his credit? '
ANSWER: str = 'According to [link_id=1] and [link_id=2], James Richard Kelly has 5 films to his credit.'
REF_ANSWER: str = 'Paul Ludwig Stein (4 February 1892 - 2 May 1951) was an Austrian-born film director with 67 films to his credit.'

# the ANSWER does not imply the REF_ANSWER because ANSWER does not mention Paul Ludwig Stein.
Check_Answer(ORIGINAL_QUESTION, ANSWER, REF_ANSWER) = False  # [END]

#########################
# Example 3:
#########################

ORIGINAL_QUESTION: str = 'Are the genuses Michelia and Stenomesson in the same family?'
ANSWER: str = 'Based on the information we have, we are not sure whether the genuses Michelia and Stenomesson are in the same family [link_id=3] does not contain this information.'
REF_ANSWER: str = 'no'

# The ANSWER did not determine whether or not Michelia and Stenomesson are in the same family.
# The REF_ANSWER implies that they are not in the same family.
# Thus we cannot infer the REF_ANSWER given the ANSWER.
Check_Answer(ORIGINAL_QUESTION, ANSWER, REF_ANSWER) = False  # [END]

#########################
# Example 4:
#########################

ORIGINAL_QUESTION: str = 'In what year was the winner of the 44th edition of the Miss World competition born?'
ANSWER: str = 'According to [link_id=2], the winner of the 44th edition of the Miss World competition is Aishwarya Rai.\nAccording to [link_id=4], she was born in 1973.'
REF_ANSWER: str = '1973'

# The ANSWER implies 1973 as the final answer to ORIGINAL_QUESTION.
# REF_ANSWER implies 1973 as the final answer to ORIGINAL_QUESTION, too.
Check_Answer(ORIGINAL_QUESTION, ANSWER, REF_ANSWER) = True  # [END]

#########################
# Example 5:
#########################

ORIGINAL_QUESTION: str = 'Skull & Bones is a mini-series by a man that is an editor of what two comics?'
ANSWER: str = 'The question is ill-formed or out-of-date. We should ask for clarification from the asker.'
REF_ANSWER: str = 'Marvel Comics and DC Comics'

# The ANSWER did not provide any answer to ORIGINAL_QUESTION.
Check_Answer(ORIGINAL_QUESTION, ANSWER, REF_ANSWER) = False  # [END]

#########################
# Example 6:
#########################
\end{lstlisting}
\end{figure}

\paragraph{Ranking "Reward" Model.} The full prompt of the reward model is shown in Listing~\ref{lst:reward_model}. Notice that this is the only prompt that is 0-shot and doesn't use "code as the prompt" approach, as it's intended to be used with an instruction-tuned model.

\begin{figure}[h]
\centering
\begin{lstlisting}[style=mypy, caption=Reward model (full prompt), label=lst:reward_model]
"""Rater Instructions:
- The goal of this rating is to filter out bad actions, so that they'll be excluded from the fine-tuning dataset.
- Overall, we want to the agent to produce relevant and grounded answers with minimal steps. Anything that deviates from this goal is considered bad.
- If any element (thoughts, comments etc.) is empty, then it's automatically bad.

#########################################

*** Model Can See:
```
{inputs}
```

*** Model Output #1:
```
{action1}
```

*** Model Output #2:
```
{action2}
```

*** Model Output #3:
```
{action3}
```

*** Model Output #4:
```
{action4}
```

#########################################

Your Instructions:
- Choose the best model output based on the rater's instructions.
- Don't assume in your decision that the model knows anything outside of "Model Can See" section.
- Be specific in your explanation.

Output 3 lines when answering and make sure to follow the precise format.

Explanation: why you think model output #X is the best
Answer: #X
Ranking: #X > #Y > ...
"""
\end{lstlisting}
\end{figure}

\subsection{Additional data}

\begin{table}[h]
\centering
\caption{Agent self-improvement and self-distillation, Bamboogle auto-eval, mean accuracy and standard deviation over 10 runs, (\%). Evaluation before self-critique steps}
\label{tab:overall-no-critique}
\begin{tabular}{l|c|c|c}
\toprule
& \texttt{XS} & \texttt{S} & \texttt{L} \\
\midrule
Pre-trained & N/A & N/A & 69.5\textsuperscript{$\pm 2.8$} (-0.8) \\
Pilot, human filtered & 44.3\textsuperscript{$\pm 3.0$} (-0.4) & 54.4\textsuperscript{$\pm 4.1$} (-2.2) & 70.9\textsuperscript{$\pm 3.0$} (-0.6) \\
Self-improvement, 1st gen & 54.8\textsuperscript{$\pm 3.7$} (+0.4) & 61.2\textsuperscript{$\pm 2.5$} (-0.7) & 73.1\textsuperscript{$\pm 3.0$} (-0.9) \\
Self-improvement, 2nd gen & 65.6\textsuperscript{$\pm 3.0$} (-0.3) & 69.2\textsuperscript{$\pm 1.8$} (-0.5) & 75.0\textsuperscript{$\pm 1.3$} (-1.1) \\
\bottomrule
\end{tabular}
\end{table}